\documentclass{Interspeech2024}

\interspeechcameraready

\title{Text Injection for Neural Contextual Biasing}

\name[affiliation={}]{Zhong}{Meng}
\name[affiliation={}]{Zelin}{Wu}
\name[affiliation={}]{Rohit}{Prabhavalkar}
\name[affiliation={}]{Cal}{Peyser}
\name[affiliation={}]{Weiran}{Wang}
\name[affiliation={}]{Nanxin}{Chen}
\name[affiliation={}]{\\Tara N.}{Sainath}
\name[affiliation={}]{Bhuvana}{Ramabhadran}

\address{
  Google LLC., USA}
\email{zhongmeng@google.com}

\keywords{ASR, text injection, contextual biasing}

\begin{document}

\maketitle

\begin{abstract}
    
    Neural contextual biasing effectively improves automatic speech recognition (ASR) for crucial phrases within a speaker's context, particularly those that are infrequent in the training data. This work proposes contextual text injection (CTI) to enhance contextual ASR. CTI leverages not only the paired speech-text data, but also a much larger corpus of unpaired text to optimize the ASR model and its neural biasing component.
    Unpaired text is converted into speech-like representations and used to guide the model's attention towards relevant bias phrases.
    Moreover, we introduce a contextual text-injected (CTI) minimum word error rate (MWER) training, which minimizes the expected WER caused by contextual biasing when unpaired text is injected into the model. 
    Experiments show that CTI with 100 billion text sentences can achieve up to 43.3\% relative WER reduction from a strong neural biasing model. CTI-MWER provides a further relative improvement of 23.5\%.
\end{abstract}

\section{Introduction}
\label{sec:introduction}
End-to-end (E2E) models excel in automatic speech recognition (ASR) by directly mapping speech signals to word sequences \cite{chiu2018state, jain2019rnn, sainath2020streaming, li2020developing, zeyer2020new}.
Despite training on large audio-transcript datasets, E2E ASR on contextually relevant words that appear infrequently in the training data remains challenging \cite{sainath2021efficient}. Such phrases include user's contact names, songs, applications, and current locations, etc.
Contextual biasing offers an effective solution to mitigate this training-inference mismatch, improving ASR performance on contextual long-tail words.

Traditional contextual biasing approaches train an external language model (LM) \cite{vasserman2016contextual} or construct a contextual finite state transducer (FST) \cite{aleksic2015interspeech, mcgraw2016personalized, zhao2019shallow, wang2023contextual}, and combine their scores with the ASR model during inference. These methods separately optimizes the ASR model and biasing component, requiring extensive tuning for the interpolation weights. E2E neural biasing \cite{pundak2018deep, jain2020contextual, chang2021context, munkhdalai2022fast} addresses this by directly integrating the biasing component into the ASR model, enabling joint optimization towards a single ASR objective.
These methods employ attention mechanism to associate bias phrases with the decoder or the acoustic encoder of the ASR model, generating a context vector that adapts the ASR model towards the biasing context. Neural biasing shows improved ASR performance compared to the FST-based approach.
However, these methods face a scalability challenge: as the number of biasing phrases increases, both the computational latency and word error rate (WER) tend to rise.
To address this, neural associate memory (NAM) \cite{munkhdalai2023nam+, wu2023dual} utilizes a two-pass hierarchical biasing scheme: the 1st pass identifies the top-K bias phrases that are most likely to occur in the speech signal via correlation, and the 2nd pass restricts the acoustic encoder's attention into only the selected K phrases.
Deferred NAM \cite{wu2023deferred} further improves the 1st pass top-K search by introducing a retrieval loss, and has achieved the state-of-the-art (SOTA) performance on multi-context biasing sets.


However, training neural biasing models requires paired audio-transcript data, which is limited by the cost of human transcription. Moreover, bias phrases sampled from such transcripts are neither rare phrases for the model nor contextually relevant to the test utterances.
To address this, we propose a novel \emph{contextual text injection (CTI)}. 
CTI transforms large-scale \emph{unpaired text} into speech-like representations through an ASR acoustic encoder, and associates them with bias phrases extracted from the \emph{unpaired text} via an attention mechanism. This enhances the model's ability to identify the most relevant bias phrases and bias its predictions towards them. CTI offers significant advantages: 1) unpaired text is readily accessible, and is far more plentiful than audio-transcript pairs; 2) unpaired text contains a wealth of rare phrases that are infrequent or even absent in audio-transcript pairs. 
Further, we propose contextual text-injected (CTI) minimum word error rate (MWER) training, where unpaired text, after undergoing contextual neural biasing, is used to generate N-best hypotheses. 
Minimizing an extra text-based MWER loss computed from such hypotheses effectively reduces ASR errors introduced by contextual biasing.
Experimented with 100 billion (B) text sentences, CTI reduces WER by up to 43.3\% relatively compared to a SOTA deferred NAM on voice search multi-context biasing sets. 
CTI MWER leads to an additional 23.5\% relative WER reduction from CTI.



\section{Related Work}
\subsection{Text Injection}
Various studies have explored incorporating unpaired text into E2E model training. 
One approach is to fine-tune the decoder of an E2E model with unpaired text to minimize an additional cross-entropy LM loss \cite{pylkkonen2021fast, meng2021ilma, chen2022factorized, meng2022modular, meng2023jeit, bijwadia2023text, gong2024advanced}. 
By zeroing out the encoder output, the decoder acts as an internal LM \cite{variani2020hybrid, meng2021ilme, meng2021ilmt}. However, the decoder's limited capability restricts the effectiveness of these methods.
Text injection \cite{bapna2021slam,tang2022unified,thomas2022towards,chen2022maestro,sainath2022joist,peyser2023improving, wei2023conversational} addresses this by directly feeding unpaired text into the ASR acoustic encoder, where the text is mapped to a shared latent space with speech, either explicitly or implicitly. The latent embeddings are then used to train the E2E models.

Text injection is initially applied to contextual biasing in \cite{sainath2023improving}. Building upon JOIST \cite{sainath2022joist}, \cite{sainath2023improving} injects contextually relevant, unpaired text directly 
into an ASR model that does \emph{not} contain any biasing component. This improves performance when combined with an FST-based context model during inference.
Our CTI method differs by applying JOIST-like text injection to a neural biasing model. This allows joint optimization of ASR model and its biasing component using unpaired text towards text-based ASR and retrieval objectives. Consequently, CTI integrates the advantages of both powerful neural biasing and text injection, while eliminating the need for FST or any other external biasing component during inference. 

\subsection{Neural Contextual Biasing with Deferred NAM}
\label{sec:nam}
Upon removing red lines and red boxes, Fig. \ref{fig:cti} depicts a speech transducer \cite{graves2012sequence, variani2020hybrid} with a neural biaser, deferred NAM \cite{wu2023deferred}.
The transducer consists of an acoustic encoder, a decoder and a joint network. The acoustic encoder transforms input speech features $\mathbf{X}$ into acoustic embeddings.
To recognize named entities in speaker's context, 
each speech utterance $\mathbf{X}$ is assigned $N$ bias phrases $Z=\{z_1,..., z_N\}$. $z_n$ may or may not exist in $\mathbf{Y}$. Deferred NAM performs 2-pass hierarchical biasing at layer $l$ of the acoustic encoder.
In Pass 1, we retrieve the K most relevant biasing phrases based on the correlations between acoustic embeddings $\mathbf{H}_l$ and bias phrase embeddings $\mathbf{P}$, followed by a max-pooling over $T$:
\begin{align}
\mathbf{a} = \text{MaxPool}_{t=1}^T \left(\mathbf{h}_{l, t} {\mathbf{P}}^\top / \sqrt{d} \right), \label{eqn:phrase_correlation}
\end{align}
where $\mathbf{a} \in \mathbb{R}^{N+1}$, $\mathbf{H}_l = \{\mathbf{h}_{l, 1}, \ldots, \mathbf{h}_{l, T}\}, \mathbf{h}_{l,t} \in \mathbb{R}^{d}$ is the sequence of acoustic embeddings at the biasing layer $l$. $\mathbf{P} \in \mathbb{R}^{(N + 1)\times d}$ is generated by feeding $Z$ through a phrase encoder and appending the output with a $d$-dim NO\_BIAS embedding. \footnote{For simplicity, we assume all encoder embeddings in this work have the same dimension $d$ and omit projection matrices.}
We select top-K phrases from $Z$ corresponding to the K largest values of the correlation vector $\mathbf{a}$, and tokenize them into word-pieces $Z^\text{w}=\{z^\text{w}_1, \ldots, z^\text{w}_M\}$, $M\ge K$. 

Then in Pass 2, we compute the cross-attention between acoustic embeddings $\mathbf{H}_l$ (as queries) and the word-piece embeddings
$\mathbf{W} \in \mathbb{R}^{M\times d}$ of $Z^\text{w}$ (as keys and values
\footnote{For better biasing performance, values are typically obtained by left-shifting keys by one token and appending a zero-embedding \cite{munkhdalai2022fast}.}):
\begin{align}
\mathbf{C} = \text{CrossAttention}(Q=\mathbf{H}_l; K,V=\mathbf{W}) \label{eqn:cross_attn},
\end{align}
where $\mathbf{C} \in \mathbb{R}^{T\times d}$ contains the $T$ context vectors generated by the cross-attention $\text{Softmax}(QK^\top / \sqrt{d})V$.

The \emph{biased} acoustic embeddings are obtained by $\mathbf{H}^\text{bias}_l = \mathbf{H}_l + \lambda_c\mathbf{C}, \lambda_c \in \mathbb{R}$,
which are then propagated through the upper acoustic encoder layers. The decoder takes in previous labels to generate the current label embedding. 
The joint network combines the top-layer acoustic and label embeddings
 via a feed-forward network, and estimates posteriors of the ground-truth transcript $\mathbf{Y}$. 
A transducer loss $\mathcal{L}^\text{ASR}(\mathcal{D}) = -\sum_{(\mathbf{X}, \mathbf{Y}, Z)\in \mathcal{D}}\log P(\mathbf{Y}|\mathbf{X}, Z)$ is calculated on supervised audio-transcript data $\mathcal{D}$.
In addition, to ensure the K phrases selected in Pass 1 are most relevant to $\mathbf{X}$, we minimize a retrieval loss 
$\mathcal{L}^\text{retr}$ computed by a cross-entropy between
$\mathbf{a}$ and the ground-truth label $r$.
$r$ indicates the longest bias phrase present in $\mathbf{Y}$, or NO\_BIAS if none exists. 
Likewise, we calculate a word-piece retrieval loss based on the correlations between $\mathbf{H}_l$ and $\mathbf{W}$,
%
and add it to $\mathcal{L}^\text{retr}(\mathcal{D})$.
We jointly minimize the transducer loss and the retrieval loss $\lambda_a\mathcal{L}^\text{ASR}(\mathcal{D}) + \lambda_d \mathcal{L}^\text{retr}(\mathcal{D})$, $\lambda_a, \lambda_d \in \mathbb{R}$ are loss weights.

\section{Contextual Text Injection}
\label{sec:text_inject_biasing}

In this work, we propose \emph{contextual text injection (CTI)}, in which a large corpus of unpaired text $\mathbf{Y}^\text{text} \in \mathcal{D}^\text{text}$ together with its contextual bias phrases $Z^\text{text}$ are injected into an ASR model and its neural biasing component to enhance its biasing capability. 
\begin{figure}[htpb!]
    \vspace{-5pt}
	\centering
	\includegraphics[width=1.0\columnwidth]{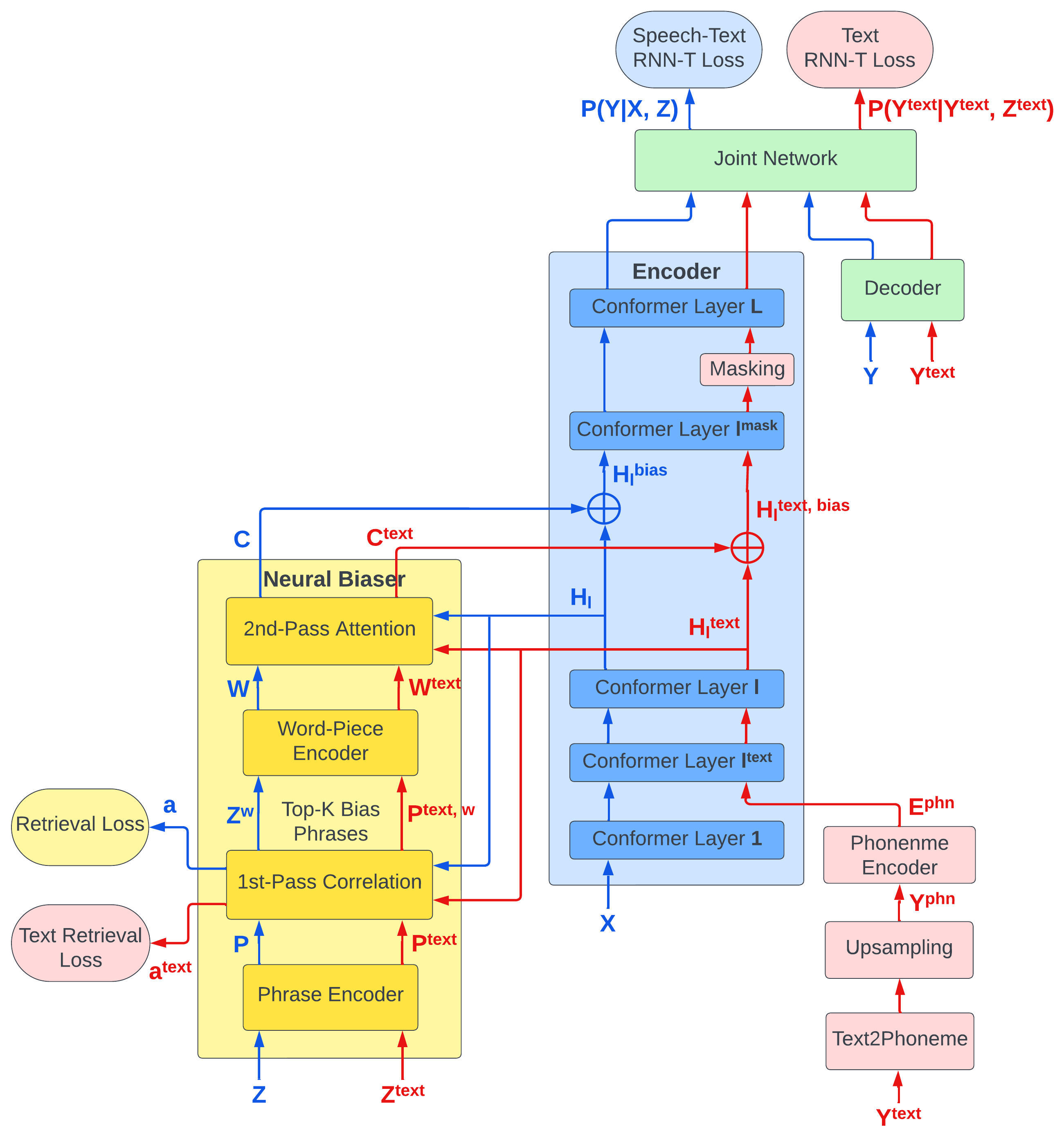}
    \vspace{-15pt}
    \caption{Contextual text injection (CTI) for a speech transducer with neural biasing. Blue and red arrows show the forward propagation paths for paired audio-transcript data and unpaired text, respectively. }
	\label{fig:cti}
 \vspace{-5pt}
\end{figure}
As shown in Fig. \ref{fig:cti}, we first tokenize unpaired text into phoneme sequences and then replicate each token a fixed or random number of times to get an upsampled sequence $\mathbf{Y}^\text{phn}=\{y^\text{phn}_1, \ldots, y^\text{phn}_I\}$.
$\mathbf{Y}^\text{phn}$ is then fed into a phoneme encoder to generate phoneme embeddings $\mathbf{E}^\text{phn} \in \mathbb{R}^{I\times d}$.
To utilize unpaired text for neural biasing,
the phoneme embeddings need to be injected to an acoustic encoder layer $l^\text{text}$ that precedes the biasing layer $l$, i.e., $l^\text{text} < l$. Therefore, different from JOIST \cite{sainath2022joist}, the masking in CTI has to be performed after the biasing layer instead of before the phoneme encoder.
We propagate $\mathbf{E}^\text{phn}$ through layer $l^\text{text}$ to layer $l$ of the acoustic encoder to generate \emph{text-based} acoustic embeddings $\mathbf{H}^\text{text}_l \in \mathbb{R}^{I\times d}$.

Each text sentence $\mathbf{Y}^\text{text}$ is associated with $N$ bias phrases $Z^\text{text} = \{z^\text{text}_1, \ldots, z^\text{text}_N\}$, sampled from sentences within $\mathbf{Y}^\text{text}$'s training mini-batch. $z^\text{text}_n$ may or may not exist in $\mathbf{Y}^\text{text}$.
$\mathbf{H}^\text{text}_l$ and  $Z^\text{text}$ undergo the two-pass contextual biasing procedure similar to Section \ref{sec:nam}.
In Pass 1, correlations between $\mathbf{H}^\text{text}_l$ and $Z^\text{text}$ are used to select the top-K bias phrases for each text sentence and to compute a retrieval loss. In Pass 2, cross-attention between $\mathbf{H}^\text{text}_l$ and only the selected bias phrases is computed to generate \emph{text-based} biased acoustic embeddings $\mathbf{H}^\text{text, bias}_l$.

Specifically, in Pass 1, we generate phrase embeddings $\mathbf{P}^\text{text}\in \mathbb{R}^{(N+1)\times d}$ by feeding $Z^\text{text}$ into the phrase encoder and appending a $d$-dim NO\_BIAS embedding to the output. We then calculate $\mathbf{a}^\text{text} \in \mathbb{R}^{N+1}$ by max-pooling the correlations between $\mathbf{H}^\text{text}_l$ and $\mathbf{P}^\text{text}$ over $I$:
\begin{align}
\mathbf{a}^\text{text} = \text{MaxPool}_{i=1}^{I} \left(\mathbf{h}^\text{text}_{l, i} {\mathbf{P}^\text{text}}^\top / \sqrt{d}\right), \label{eqn:text_phrase_correlation}
\end{align}
where $\mathbf{a}^\text{text}$ signifies the correlations between $\mathbf{H}^\text{text}_l$ and each bias phrase (and NO\_BIAS) in $Z$. 
From set $Z^\text{text}$, we select the top-K bias phrases that have the K largest correlation values in $\mathbf{a}^\text{text}$, and tokenize them into word-pieces $Z^\text{text, w} = \{z^\text{text, w}_1, \ldots, z^\text{text, w}_M\}$, $M \ge K$. 
In Pass 2, $\mathbf{H}^\text{text}_l$ first cross-attends into the word-piece embeddings $\mathbf{W}^\text{text}\in \mathbb{R}^{M\times d}$ of $Z^\text{text, w}$ to compute text-based context vectors $\mathbf{C}^\text{text}\in \mathbb{R}^{I\times d}$. $\mathbf{H}^\text{text, bias}_l\in \mathbb{R}^{I\times d}$ is then obtained by adding $\mathbf{C}^\text{text}$ to $\mathbf{H}^\text{text}_l$:
\begin{align}
& \mathbf{C}^\text{text} = \text{CrossAttention}(Q=\mathbf{H}^\text{text}_l; K, V=\mathbf{W}^\text{text}) \label{eqn:text_cross_attn}, \\
& \mathbf{H}^\text{text, bias}_l = \mathbf{H}^\text{text}_l + \lambda_c \mathbf{C}^\text{text}, \label{eqn:text_h_bias}
\end{align}
where $\lambda_c \in \mathbb{R}$ is the biasing strength.

To make the task sufficiently difficult for the model with text input,
we mask a portion of text-based biased acoustic embeddings $\mathbf{H}^\text{text, bias}_{l^\text{mask}}$ at layer $l^\text{mask}$
after the biasing layer $l$, i.e., $l^\text{mask} > l$, by following \cite{baevski2020wav2vec}. 
The masked embeddings are then propagated through the rest of the transducer,
contributing to the calculation of a text-based transducer loss:
\begin{align}
  \mathcal{L}^\text{ASR}(\mathcal{D}^\text{text}) = - \log P(\mathbf{Y}^\text{text}|\text{Mask}(\mathbf{H}^\text{text, bias}_{l^\text{mask}}), Z^\text{text}).
\end{align}
To accurately identify the K most relevant bias phrases from the unpaired text, we minimize a retrieval loss $\mathcal{L}^\text{retr}(\mathcal{D}^\text{text})$ computed as a cross-entropy between the correlation $\mathbf{a}^\text{text}$ and ground-truth label $r^\text{text}$. 
$r^\text{text}$ is assigned the longest bias phrase from $Z^\text{text}$ that appears in $\mathbf{Y}^\text{text}$. Otherwise, $r^\text{text}$ is assigned NO\_BIAS if none of the bias phrases are present in $\mathbf{Y}^\text{text}$.
Similarly,
a word-piece retrieval loss based on the same unpaired text is computed and then added to $\mathcal{L}^\text{retr}(\mathcal{D}^\text{text})$.

The speech transducer, phoneme encoder, and neural biaser are jointly trained
from scratch to minimize two ASR losses and two retrieval losses using audio-transcript paired data $\mathcal{D}$ and unpaired text $\mathcal{D}^\text{text}$:
\begin{align}
    \mathcal{L}^\text{CTI} = & \lambda_a \mathcal{L}^\text{ASR}(\mathcal{D}) + \lambda_d \mathcal{L}^\text{retr}(\mathcal{D}) \nonumber \\
    & \quad + \lambda^\text{text}_a \mathcal{L}^\text{ASR}(\mathcal{D}^\text{text}) + \lambda^\text{text}_d \mathcal{L}^\text{retr}(\mathcal{D}^\text{text}).
\end{align}
where $\lambda^\text{text}_a, \lambda^\text{text}_d \in \mathbb{R}$ are weights for the text-based losses.
With CTI, the model can learn from significantly larger pool of unpaired text the ability to extract relevant bias phrases and improve the accuracy of named entity recognition based on them.

\section{Contextual Text-Injected MWER}
\label{sec:mwer}


Conventional MWER training minimizes the expected number of word errors on supervised training data \cite{prabhavalkar2018minimum, lu2020minimum, meng2021minimum}, optionally including unpaired text \cite{sainath2022joist}. However, these methods reply on N-best hypotheses produced by a non-contextualized ASR system, limiting its ability to correct errors caused by contextual biasing.
To overcome this, we propose \emph{contextual text-injected (CTI) MWER} training. This approach generates N-best hypotheses by feeding both audio-transcript pairs $\mathcal{D}$ and unpaired text $\mathcal{D}^\text{text}$ through an ASR model with neural biasing, then computes MWER losses based on these hypotheses.

With CTI MWER, we first generate the N-best hypotheses $\{\mathbf{Y}_1, \ldots, \mathbf{Y}_N\}$ by inputting speech $\mathbf{X}$ and the bias phrases $Z$ 
into the model and applying the two-pass neural contextual biasing outlined in Section \ref{sec:nam}. 
This aligns the N-best generation with the actual inference, empowering the MWER training to rectify errors introduced by contextual biasing.
The contextual MWER loss on supervised data is computed as the expected number of word errors on these N-best hypotheses: 
\begin{align}
    \mathcal{L}^\text{MWER}(\mathcal{D})=&\sum_{n=1}^N \frac{P(\mathbf{Y}_n | \mathbf{X}, Z)}{\sum_{n=1}^N P(\mathbf{Y}_n | \mathbf{X}, Z)} R(\mathbf{Y}_n, \mathbf{Y}), \label{eqn:mwer}
\end{align}
where $R(\mathbf{Y}_n, \mathbf{Y})$ is the number of word errors in a hypothesis $\mathbf{Y}_n$ compared to the ground-truth transcript $\mathbf{Y}$.

Further, we inject unpaired text $\mathbf{Y}^\text{text}$ and bias phrases $Z^\text{text}$ into the ASR and its neural biasing model to
calculate an text-based MWER loss:
\begin{align}
    \mathcal{L}^\text{MWER}(\mathcal{D}^\text{text})=\sum_{n=1}^N \frac{P(\mathbf{Y}^\text{text}_n | \mathbf{H}^\text{text, bias}_{l^\text{mask}})}{\sum_{n=1}^N P(\mathbf{Y}^\text{text}_n | \mathbf{H}^\text{text, bias}_{l^\text{mask}})} R(\mathbf{Y}^\text{text}_n, \mathbf{Y}^\text{text}), 
    \label{eqn:text_mwer} \nonumber
\end{align}
where $\{\mathbf{Y}^\text{text}_1, \ldots, \mathbf{Y}^\text{text}_N\}$ are N-best hypotheses generated via the text-based two-pass neural biasing in Section \ref{sec:text_inject_biasing}.
Finally, we minimize the CTI-MWER loss across both $\mathcal{D}$ and $\mathcal{D}^\text{text}$:
\begin{align}
   \mathcal{L}^\text{CTI-MWER} = \lambda_m \mathcal{L}^\text{MWER}(\mathcal{D}) + \lambda^\text{text}_m\mathcal{L}^\text{MWER}(\mathcal{D}^\text{text}),
\end{align}
where $\lambda_m, \lambda^\text{text}_m \in \mathbb{R}$ are loss weights.

To enhance the ASR performance in the \emph{absence} of bias phrases, we selectively activate the biasing component during CTI MWER training. We do this by feeding in empty bias phrases for a specific percentage of the training data. 

\section{Experiments}
\subsection{Data}
Our supervised training data comprises 520 million multi-domain English audio-transcript pairs (490 thousand (K) hours) from voice search (VS) traffic as in \cite{prabhavalkar2024extreme}.
In addition, we leverage a much larger corpus of unpaired text for text injection, exceeding the size of audio-transcript pairs by two orders of magnitude. This corpus consists of 100B anonymized sentences drawn from Maps, Google Play, Web, and YouTube domains as in \cite{sainath2022joist}.

We assess our models across various test sets. 
The head common-word test set comprises roughly 12K anonymized and hand-transcribed VS utterances with an average duration of 5.5 seconds.
For rare-word evaluation, we utilize a multi-context biasing corpus from \cite{munkhdalai2023nam+, wu2023deferred}, consisting of three sets: 
\begin{itemize}
    \item \textbf{NO\_PRE}: 1.3K utterances matching prefix-less patterns from \texttt{\$APPS}, \texttt{\$CONTACTS}, \texttt{\$SONGS} categories (i.e., ACS).
    \item \textbf{PRE}: 2.6K utterances matching prefixed patterns like ``open \texttt{\$APPS}'', ``call \texttt{\$CONTACTS}'', ``play \texttt{\$SONGS}''. 
    \item \textbf{ANTI}: 1K utterances from general voice assistant queries. 
\end{itemize}
In all three sets, each utterance is assigned 150, 300, 600, 1500, 3000 ACS bias phrases. The NO\_PRE and PRE sets evaluate \emph{in-context} performance where one bias phrase actually appears in the transcript truth, while the ANTI set assesses \emph{anti-context} performance where only distractor bias phrases are assigned.

\subsection{Modeling Architecture}
This work utilizes a non-streaming hybrid autoregressive transducer (HAT) \cite{variani2020hybrid} for ASR.
We extract 128-dim log Mel filterbanks from speech and subsample it with two 2D-convolution layers, resulting a 512-dim feature every 40 ms.
The HAT acoustic encoder is based on the architecture of Google USM \cite{zhang2023google}. 
It starts with 2 convolution layers, followed by 12 conformer layers \cite{gulati2020conformer}. Each conformer layer uses a 8-head local self-attention with 512 dimensions and a convolution kernel of size 10. Funnel pooling \cite{dai2020funnel} reduces the computation latency, resulting in a frame rate of 320ms at the encoder output \cite{wang2023massive, prabhavalkar2024extreme}.
The HAT decoder is a 640-dim $|V|^2$ embedding \cite{botros2021tied}.
The output vocabulary consists of 4096 lowercase word-pieces. HAT has 857M parameters in total.

The neural biasing is applied at 4th encoder layer of HAT ($l=4$). The phrase encoder is a 4-layer deep averaging network with
256 hidden units (788K parameters). The word-piece encoder is a 1-layer conformer with 256 hidden units (1M parameters). 
The projection layers in correlation and cross-attention have 7.3M parameters.
All biasing components are significantly smaller than the main HAT. 
The bias strength $\lambda_c$ is set to 1.0 during training and 0.6 during inference.



\subsection{Text injection for Neural Contextual Biasing}
\label{sec:exp_text_inject}



We first train a baseline HAT with deferred NAM \cite{wu2023deferred} using only the supervised data, and show its results in Table \ref{table:text_inject}. 
$\lambda_a=0.9, \lambda_r=0.1, K=32$ for all the experiments. Note that, for each biasing test set, the reported WER is the average across 5 different scenarios where the model receives varying number of bias entities.
\begin{table}[h]
\vspace{-0 pt}
\centering
\begin{tabular}[c]{l|c||c|c|c}
	\hline
	\hline
	\multirow{2}{*}{\begin{tabular}{@{}c@{}} \hspace{-1pt} Method \end{tabular}} &
	\multirow{2}{*}{\begin{tabular}{@{}c@{}} \hspace{7pt} VS \hspace{7pt} \end{tabular}} & 
        \multicolumn{2}{c|}{In-Context} & 
	\multirow{2}{*}{\begin{tabular}{@{}c@{}} \hspace{2pt} ANTI\hspace{2pt}  \end{tabular}} \\
        \hhline{~~--~}
         & & \hspace{-3.7pt} NO\_PRE \hspace{-3.7pt} & 
         \hspace{4.5pt} PRE \hspace{4.5pt} & \\
	\hline
        Sup & 4.1 & 3.0 & 2.3 & 1.8 \\
        JOIST & 4.0 & 2.2 & 2.1 & 1.9 \\
        CTI & \textbf{3.9} & \textbf{1.7} & \textbf{1.7} & \textbf{1.6} \\
	\hline
	\hline
	\end{tabular}
        \vspace{2pt}
	\caption[a]{WERs (\%) for HAT with neural contextual biasing.\footnotemark \quad The models are trained from scratch using supervised data (Sup), JOIST or contextual text injection (CTI).} 
\label{table:text_inject}
\vspace{-10 pt}
\end{table}
\footnotetext{Without contextual biasing, HAT with JOIST and text-injected MWER achieves 4.0\%, 22.1\%, 10.5\% and 1.6\% WERs on VS, NO\_PRE, PRE and ANTI, respectively.}
Then we perform JOIST training of HAT with deferred NAM by injecting unpaired text \emph{only} to the acoustic encoder of HAT. No neural biasing (Eqs. \eqref{eqn:text_phrase_correlation} -- \eqref{eqn:text_h_bias}) is performed with unpaired text. The settings are consistent with those used for JOIST on HAT:
the unpaired text is tokenized into phoneme sequences. Each token is repeated a random number of times. The random number is sampled from a uniform distribution between 1 and 2. We mask 30\% of the upsampled tokens with spans of $S=10$ and feed it into a phoneme encoder. The phoneme embeddings are finally inputted to the 3rd conformer layer of the acoustic encoder. $\lambda^\text{text}_a$ is set to 0.1 for all the experiments.
JOIST improves the in-context biasing by 26.7\% and 8.7\% relatively on PRE and NO\_PRE, but degrades the anti-context set by 5.5\%.

Further, we conduct contextual text injection as described in Section \ref{sec:text_inject_biasing}. 
We follow the same tokenization and upsampling procedure for unpaired text as used in JOIST.
Without any masking, we directly embed these upsampled phoneme tokens via a phoneme encoder and fed them into the 3rd conformer layer of the acoustic encoder ($l^\text{text}=3$). The neural biasing is then performed with both real and text-based acoustic embeddings at the output of the 5th conformer layer ($l=5$). The text-based biased acoustic embeddings are masked at the 7th conformer layer ($l^\text{mask}$=7). $\lambda^\text{text}_r$ is set to 0.1.
CTI demonstrates substantial and consistent improvement over JOIST on all biasing test sets and VS, resulting in total relative WER reductions of 43.3\%, 26.1\% on in-context sets, 11.1\% on the anti-context set, and 4.9\% on head VS from the supervised baseline.

\subsection{MWER Training for Neural Contextual Biasing}
\label{sec:exp_mwer}
We initialize all the MWER training with the CTI model which has achieved the best WERs so far.
We first perform MWER training with only supervised data to minimize the loss in Eq. \eqref{eqn:mwer}. 
\begin{table}[h]
\centering
\setlength{\tabcolsep}{4.5pt}
\begin{tabular}[c]{l|c||c|c|c}
	\hline
	\hline
	\multirow{2}{*}{\begin{tabular}{@{}c@{}} \hspace{13pt} Method \end{tabular}} &
	\multirow{2}{*}{\begin{tabular}{@{}c@{}} \hspace{7pt} VS \hspace{7pt} \end{tabular}} & 
        \multicolumn{2}{c|}{In-Context} & 
	\multirow{2}{*}{\begin{tabular}{@{}c@{}} \hspace{2pt} ANTI\hspace{2pt}  \end{tabular}} \\
        \hhline{~~--~}
         & & \hspace{-3.7pt} NO\_PRE \hspace{-3.7pt} & 
         \hspace{4.5pt} PRE \hspace{4.5pt} & \\
	\hline
        CTI & 3.9 & 1.7 & 1.7 & \textbf{1.6} \\
        + MWER & 3.9 & 1.7 & 1.5 & 1.9 \\
        + JOIST MWER \hspace{-4pt} & \textbf{3.8} & 1.5 & 1.4 & 1.9 \\
        + CTI MWER & \textbf{3.8} & \textbf{1.3} & \textbf{1.3} & 1.8 \\
	\hline
	\hline
	\end{tabular}
        \vspace{1pt}
	\caption{WERs (\%) for HAT with neural contextual biasing. The models are trained in two stages: first with contextual text injection (CTI), and then fine-tuned using MWER, JOIST-based MWER, or contextual text-injected (CTI) MWER training.}
\label{table:mwer}
\vspace{-15 pt}
\end{table}
As in Table \ref{table:mwer}, MWER significantly improves over CTI on the in-context PRE and the head VS, but degrades on the anti-context test set. We then perform JOIST-based MWER \cite{sainath2022joist} by injecting text into only the HAT encoder. During this process, the biasing modules are updated with only the supervised data. Compared to the MWER training, JOIST-based MWER dramatically reduces WERs on NO\_PRE. Finally, we perform contextual text-injected MWER (Section \ref{sec:mwer}), updating both HAT and biasing modules with supervised data and unpaired text. CTI MWER significantly outperforms JOIST-based MWER across VS and all biasing sets.
Moreover, CTI MWER offers remarkable improvement over CTI on all test sets, achieving total relative WER reductions of a 56.7\%, 43.5\% on in-context biasing sets, and 7.3\% on head VS from the supervised baseline.
Note that, to reduce the WER also on head VS, we selectively drop out the neural biasing component for 30\% of the training mini-batches.

\subsection{Scalability to Number of Bias Phrases}
We gradually increase the number of bias phrases assigned to each test utterance (0-3,000) when evaluating HAT with deferred NAM on the NO\_PRE in-context biasing set in Table \ref{table:bias_entities}.
Despite the consistent increase in WER with more bias phrases, CTI and CTI MWER remarkably maintain WER below 1.8\% even at the maximum 3000 bias phrases. Importantly, the performance trends observed in Sections \ref{sec:exp_text_inject}, \ref{sec:exp_mwer} persist across all bias phrase counts: CTI and CTI MWER outperform the baseline supervised model, with their advantage increasing as the number of phrases decreases. These trends hold consistently for PRE and ANTI biasing sets as well.

\begin{table}[h]
\centering
\setlength{\tabcolsep}{3.8pt}
\begin{tabular}[c]{l|c||c|c|c|c|c|c}
	\hline
	\hline
	\hspace{0pt}\multirow{2}{*}{\begin{tabular}{@{}c@{}} \hspace{10pt} Method
        \end{tabular}\hspace{0pt}} & VS & 
        \multicolumn{6}{c}{In-Context: NO\_PRE} \\
        \hhline{~-------}
        & \hspace{6pt}0\hspace{6pt} & 0 & \hspace{1pt}150\hspace{1pt} & \hspace{1pt}300\hspace{1pt} & \hspace{1pt}600\hspace{1pt} & \hspace{-1.7pt}1500\hspace{-1.7pt} & \hspace{-1.7pt}3000\hspace{-1.7pt} \\
	\hline
        Sup & 4.1 & \hspace{-0.1pt}22.2\hspace{-0.1pt} & 2.7 & 2.7 & 2.8 & 3.4 & 3.5 \\
	\hline
        CTI & 3.9 & \textbf{21.8} & 1.3 & 1.4 & 1.7 & 1.8 & 2.1 \\
        + CTI MWER & \textbf{3.8} & \textbf{21.8} & \textbf{1.1} & \textbf{1.2} & \textbf{1.2} & \textbf{1.4} & \textbf{1.8} \\
        \hline
	\hline
	\end{tabular}
        \vspace{1pt}
	\caption{WERs (\%) for supervised training, contextual text injection (CTI) and contextual text-injected (CTI) MWER training with respect to the number of bias phrases (0-3000) assigned to each test utterance.}
    \vspace{-25pt}
\label{table:bias_entities}
\end{table}


\section{Conclusion}
We propose a novel contextual text injection that injects unpaired text to not only the ASR encoder, but also its biasing component. We further introduce contextual text-injected MWER training, a method that fine-tunes the ASR model with neural biasing to minimize an additional text-based MWER objective. CTI reduces the WER of a HAT with deferred NAM by 26.1\%-43.3\% relatively on in-context biasing sets, 11.1\% relatively on the anti-context set, and 4.9\% on voice search.  Further, CTI MWER achieves additional 23.5\% relative WER reductions over CTI on in-context biasing sets.

\newpage

\bibliographystyle{IEEEtran}
\bibliography{mybib}

\begin{thebibliography}{10}
\providecommand{\url}[1]{#1}
\csname url@samestyle\endcsname
\providecommand{\newblock}{\relax}
\providecommand{\bibinfo}[2]{#2}
\providecommand{\BIBentrySTDinterwordspacing}{\spaceskip=0pt\relax}
\providecommand{\BIBentryALTinterwordstretchfactor}{4}
\providecommand{\BIBentryALTinterwordspacing}{\spaceskip=\fontdimen2\font plus
\BIBentryALTinterwordstretchfactor\fontdimen3\font minus \fontdimen4\font\relax}
\providecommand{\BIBforeignlanguage}[2]{{%
\expandafter\ifx\csname l@#1\endcsname\relax
\typeout{** WARNING: IEEEtran.bst: No hyphenation pattern has been}%
\typeout{** loaded for the language `#1'. Using the pattern for}%
\typeout{** the default language instead.}%
\else
\language=\csname l@#1\endcsname
\fi
#2}}
\providecommand{\BIBdecl}{\relax}
\BIBdecl

\bibitem{chiu2018state}
C.-C. Chiu, T.~N. Sainath, and Y.~Wu, ``State-of-the-art speech recognition with sequence-to-sequence models,'' in \emph{Proc. ICASSP}, 2018.

\bibitem{jain2019rnn}
M.~Jain, K.~Schubert, J.~Mahadeokar \emph{et~al.}, ``{RNN-T} for latency controlled {ASR} with improved beam search,'' \emph{arXiv preprint arXiv:1911.01629}, 2019.

\bibitem{sainath2020streaming}
T.~Sainath, Y.~He, B.~Li \emph{et~al.}, ``A streaming on-device end-to-end model surpassing server-side conventional model quality and latency,'' in \emph{Proc. ICASSP}, 2020.

\bibitem{li2020developing}
J.~Li, R.~Zhao, Z.~Meng \emph{et~al.}, ``Developing {RNN-T} models surpassing high-performance hybrid models with customization capability,'' in \emph{Interspeech}, 2020.

\bibitem{zeyer2020new}
A.~Zeyer, A.~Merboldt, R.~Schl{\"u}ter \emph{et~al.}, ``A new training pipeline for an improved neural transducer,'' \emph{Proc. Interspeech}, 2020.

\bibitem{sainath2021efficient}
T.~Sainath, Y.~He, A.~Narayanan \emph{et~al.}, ``An efficient streaming non-recurrent on-device end-to-end model with improvements to rare-word modeling,'' \emph{Proc. Interspeech}, 2021.

\bibitem{vasserman2016contextual}
L.~Vasserman, B.~Haynor, and P.~Aleksic, ``Contextual language model adaptation using dynamic classes,'' in \emph{Proc. SLT}.\hskip 1em plus 0.5em minus 0.4em\relax IEEE, 2016.

\bibitem{aleksic2015interspeech}
P.~Aleksic, M.~Ghodsi, A.~Michaely \emph{et~al.}, ``{Bringing contextual information to google speech recognition},'' in \emph{Proc. Interspeech}, 2015.

\bibitem{mcgraw2016personalized}
I.~McGraw, R.~Prabhavalkar, R.~Alvarez \emph{et~al.}, ``Personalized speech recognition on mobile devices,'' in \emph{Proc. ICASSP}.\hskip 1em plus 0.5em minus 0.4em\relax IEEE, 2016.

\bibitem{zhao2019shallow}
D.~Zhao, T.~Sainath, D.~Rybach \emph{et~al.}, ``Shallow-fusion end-to-end contextual biasing.'' in \emph{Interspeech}, 2019, pp. 1418--1422.

\bibitem{wang2023contextual}
W.~Wang, Z.~Wu, D.~Caseiro \emph{et~al.}, ``Contextual biasing with the {K}nuth-{M}orris-{P}ratt matching algorithm,'' \emph{arXiv preprint arXiv:2310.00178}, 2023.

\bibitem{pundak2018deep}
G.~Pundak, T.~N. Sainath, R.~Prabhavalkar \emph{et~al.}, ``Deep context: end-to-end contextual speech recognition,'' in \emph{Proc. SLT}.\hskip 1em plus 0.5em minus 0.4em\relax IEEE, 2018.

\bibitem{jain2020contextual}
M.~Jain, G.~Keren, J.~Mahadeokar \emph{et~al.}, ``Contextual {RNN-T} for open domain asr,'' \emph{Proc. Interspeech}, 2020.

\bibitem{chang2021context}
F.-J. Chang, J.~Liu, M.~Radfar \emph{et~al.}, ``Context-aware transformer transducer for speech recognition,'' in \emph{Proc. ASRU}.\hskip 1em plus 0.5em minus 0.4em\relax IEEE, 2021.

\bibitem{munkhdalai2022fast}
T.~Munkhdalai, K.~C. Sim, A.~Chandorkar \emph{et~al.}, ``Fast contextual adaptation with neural associative memory for on-device personalized speech recognition,'' in \emph{Proc. ICASSP}.\hskip 1em plus 0.5em minus 0.4em\relax IEEE, 2022.

\bibitem{munkhdalai2023nam+}
T.~Munkhdalai, Z.~Wu, G.~Pundak \emph{et~al.}, ``{NAM}+: Towards scalable end-to-end contextual biasing for adaptive asr,'' in \emph{Proc. SLT}.\hskip 1em plus 0.5em minus 0.4em\relax IEEE, 2023.

\bibitem{wu2023dual}
Z.~Wu, T.~Munkhdalai, P.~Rondon \emph{et~al.}, ``Dual-mode {NAM}: Effective top-k context injection for end-to-end asr,'' \emph{Proc. Interspeech}, 2023.

\bibitem{wu2023deferred}
Z.~Wu, G.~Song, C.~Li \emph{et~al.}, ``Deferred {NAM}: Low-latency top-{K} context injection via deferred context encoding for non-streaming {ASR},'' \emph{Proc. NAACL 2024 - Industry Track}, 2024.

\bibitem{pylkkonen2021fast}
J.~Pylkk{\"o}nen, A.~Ukkonen, J.~Kilpikoski \emph{et~al.}, ``Fast text-only domain adaptation of {RNN}-transducer prediction network,'' \emph{Proc. Interspeech}, 2021.

\bibitem{meng2021ilma}
Z.~Meng, Y.~Gaur, N.~Kanda \emph{et~al.}, ``Internal language model adaptation with text-only data for end-to-end speech recognition,'' in \emph{Interspeech}, 2022.

\bibitem{chen2022factorized}
X.~Chen, Z.~Meng \emph{et~al.}, ``Factorized neural transducer for efficient language model adaptation,'' in \emph{Proc. ICASSP}, 2022.

\bibitem{meng2022modular}
Z.~Meng, T.~Chen, R.~Prabhavalkar \emph{et~al.}, ``Modular hybrid autoregressive transducer,'' in \emph{Proc. SLT}.\hskip 1em plus 0.5em minus 0.4em\relax IEEE, 2022.

\bibitem{meng2023jeit}
Z.~Meng, W.~Wang, R.~Prabhavalkar \emph{et~al.}, ``{JEIT}: Joint end-to-end model and internal language model training for speech recognition,'' in \emph{Proc. ICASSP}.\hskip 1em plus 0.5em minus 0.4em\relax IEEE, 2023.

\bibitem{bijwadia2023text}
S.~Bijwadia, S.-y. Chang, W.~Wang \emph{et~al.}, ``Text injection for capitalization and turn-taking prediction in speech models,'' \emph{Proc. Interspeech}, 2023.

\bibitem{gong2024advanced}
X.~Gong, Y.~Wu, J.~Li \emph{et~al.}, ``Advanced long-content speech recognition with factorized neural transducer,'' \emph{TASLP}, 2024.

\bibitem{variani2020hybrid}
E.~Variani, D.~Rybach, C.~Allauzen \emph{et~al.}, ``Hybrid autoregressive transducer ({HAT}),'' in \emph{Proc. ICASSP}.\hskip 1em plus 0.5em minus 0.4em\relax IEEE, 2020.

\bibitem{meng2021ilme}
Z.~Meng, S.~Parthasarathy, E.~Sun \emph{et~al.}, ``Internal language model estimation for domain-adaptive end-to-end speech recognition,'' in \emph{Proc. SLT}.\hskip 1em plus 0.5em minus 0.4em\relax IEEE, 2021.

\bibitem{meng2021ilmt}
Z.~Meng, N.~Kanda, Y.~Gaur \emph{et~al.}, ``Internal language model training for domain-adaptive end-to-end speech recognition,'' in \emph{Proc. ICASSP}.\hskip 1em plus 0.5em minus 0.4em\relax IEEE, 2021.

\bibitem{bapna2021slam}
A.~Bapna, Y.-A. Chung, N.~Wu \emph{et~al.}, ``{SLAM: A unified encoder for speech and language modeling via speech-text joint pre-training},'' \emph{arXiv preprint arXiv:2110.10329}, 2021.

\bibitem{tang2022unified}
Y.~Tang, H.~Gong, N.~Dong \emph{et~al.}, ``{Unified speech-text pre-training for speech translation and recognition},'' in \emph{Proc. ACL}, 2022.

\bibitem{thomas2022towards}
S.~Thomas, H.~J. Kuo, B.~Kingsbury \emph{et~al.}, ``{Towards reducing the need for speech training data to build spoken language understanding systems},'' in \emph{Proc. ICASSP}, 2022.

\bibitem{chen2022maestro}
Z.~Chen, Y.~Zhang, A.~Rosenberg \emph{et~al.}, ``{MAESTRO: Matched speech text representations through modality matching},'' in \emph{Proc. Interspeech}, 2022.

\bibitem{sainath2022joist}
T.~Sainath, R.~Prabhavalkar \emph{et~al.}, ``{JOIST}: A joint speech and text streaming model for {ASR},'' \emph{Proc. SLT}, 2022.

\bibitem{peyser2023improving}
C.~Peyser, Z.~Meng, K.~Hu \emph{et~al.}, ``Improving joint speech-text representations without alignment,'' \emph{Proc. Interspeech}, 2023.

\bibitem{wei2023conversational}
K.~Wei, B.~Li, H.~Lv \emph{et~al.}, ``Conversational speech recognition by learning audio-textual cross-modal contextual representation,'' \emph{arXiv preprint arXiv:2310.14278}, 2023.

\bibitem{sainath2023improving}
T.~N. Sainath, R.~Prabhavalkar, D.~Caseiro \emph{et~al.}, ``Improving contextual biasing with text injection,'' in \emph{Proc. ICASSP}.\hskip 1em plus 0.5em minus 0.4em\relax IEEE, 2023.

\bibitem{graves2012sequence}
A.~Graves, ``Sequence transduction with recurrent neural networks,'' \emph{Proc. ICML}, 2012.

\bibitem{baevski2020wav2vec}
A.~Baevski, Y.~Zhou, A.~Mohamed \emph{et~al.}, ``wav2vec 2.0: A framework for self-supervised learning of speech representations,'' \emph{Proc. NeurlIPS}, 2020.

\bibitem{prabhavalkar2018minimum}
R.~Prabhavalkar, T.~Sainath, Y.~Wu \emph{et~al.}, ``Minimum word error rate training for attention-based sequence-to-sequence models,'' in \emph{Proc. ICASSP}.\hskip 1em plus 0.5em minus 0.4em\relax IEEE, 2018.

\bibitem{lu2020minimum}
L.~Lu, Z.~Meng, N.~Kanda \emph{et~al.}, ``On minimum word error rate training of the hybrid autoregressive transducer,'' \emph{arXiv preprint arXiv:2010.12673}, 2020.

\bibitem{meng2021minimum}
Z.~Meng, Y.~Wu, N.~Kanda \emph{et~al.}, ``Minimum word error rate training with language model fusion for end-to-end speech recognition,'' \emph{Proc. Interspeech}, 2021.

\bibitem{prabhavalkar2024extreme}
R.~Prabhavalkar, Z.~Meng, W.~Wang \emph{et~al.}, ``Extreme encoder output frame rate reduction: Improving computational latencies of large end-to-end models,'' \emph{Proc. ICASSP}, 2024.

\bibitem{zhang2023google}
Y.~Zhang, W.~Han, J.~Qin \emph{et~al.}, ``Google {USM}: Scaling automatic speech recognition beyond 100 languages,'' \emph{arXiv preprint arXiv:2303.01037}, 2023.

\bibitem{gulati2020conformer}
A.~Gulati, J.~Qin, C.~Chiu \emph{et~al.}, ``Conformer: Convolution-augmented transformer for speech recognition,'' \emph{Proc. Interspeech}, 2020.

\bibitem{dai2020funnel}
Z.~Dai, G.~Lai, Y.~Yang \emph{et~al.}, ``Funnel-transformer: Filtering out sequential redundancy for efficient language processing,'' \emph{Proc. NeurlIPS}, 2020.

\bibitem{wang2023massive}
W.~Wang, R.~Prabhavalkar, D.~Hwang \emph{et~al.}, ``Massive end-to-end models for short search queries,'' \emph{arXiv preprint arXiv:2309.12963}, 2023.

\bibitem{botros2021tied}
R.~Botros, T.~N. Sainath, R.~David \emph{et~al.}, ``Tied \& reduced {RNN-T} decoder,'' \emph{Proc. Interspeech}, 2021.

\end{thebibliography}

\end{document}